# Defective samples simulation through Neural Style Transfer for automatic surface defect segment

Taoran Wei, Danhua Cao*, Xingru Jiang, Caiyun Zheng, Lizhe Liu

School of Optical and Electronic Information, Huazhong University of Science and Technology, Wuhan, China 430074

## ABSTRACT

Owing to the lack of defect samples in industrial product quality inspection, trained segmentation model tends to overfit when applied online. To address this problem, we propose a defect sample simulation algorithm based on neural style transfer. The simulation algorithm requires only a small number of defect samples for training, and can efficiently generate simulation samples for next-step segmentation task. In our work, we introduce a masked histogram matching module to maintain color consistency of the generated area and the true defect. To preserve the texture consistency with the surrounding pixels, we take the fast style transfer algorithm to blend the generated area into the background. At the same time, we also use the histogram loss to further improve the quality of the generated image. Besides, we propose a novel structure of segment net to make it more suitable for defect segmentation task. We train the segment net with the real defect samples and the generated simulation samples separately on the button datasets. The results show that the F1 score of the model trained with only the generated simulation samples reaches 0.80, which is better than the real sample result.

## 1. INTRODUCTION

Quantitative measurement of high-precision surface defects has always been a difficult problem in automated production. At present, a large number of related works use conventional image processing techniques[1–3] or neural network[4–6] for defect segmentation. Traditional methods lack generalization and are only valid for a specific type of defect in a particular dataset[1]. While the detection algorithm based on the neural network can be applied to a variety of defects to achieve higher detection accuracy[4]. But training a test network requires a lot of labeled images.

For specific detection objects, defect-free samples are easy to obtain, while the defect samples are not. How to make full use of the rare defect samples for online detection has become a bottleneck of automation in some fields. At present, the common solutions are mainly divided into two types: 1. Use a small network such as U-net to train several layers with a small number of images, similar to medical image segmentation. However, these methods do not achieve a satisfying effect for small defects of various shapes. 2. Expand the defect sample using methods such as rotation, cropping, and radiation. These methods do not work well for extreme cases (only a single image of each type of defect) and the trained model is prone to overfitting.

In this paper, our research object is button, an industrial product with complex texture background. There are nine types of defects on button, including stain, scratch, and hole. Defect samples are difficult to collect, only take up 3 percent of the total.

To solve this problem, we propose a novel defect segmentation framework based only on a few real defect samples, which includes simulation samples generation and defect segmentation. The main contributions are as follows:

1    We design a defect sample generation technique based on histogram matching and style transfer, named Defect Style Transfer (DST). The algorithm can generate simulation defect samples to expand the training dataset of segmentation. By learning the color information of a single reference sample, the trained model can generate the same type of defect in the specified area of the defect-free sample. Training with the generated simulation defect samples can improve the accuracy of the defect segmentation network.

*dhcao@hust.edu.cn; phone 027-87544833;

2. According to the characteristics of the defect segmentation task, we design a novel defect segmentation network named Buttonlab. In view of the fact that the sample background is relatively simple and the defect characteristics are not much different in the defect detection task, we remove the ASPP module in deeplab and add a branch network to extract local information. When trained with the same samples, Buttonlab can achieve a 0.11~0.18 higher F1 score compared to Deeplab v3+ algorithm.

3. Through the combination of the above two algorithms, on the button dataset, our proposed algorithms can achieve better results in the case that the defect samples are insufficient. The F1 score of Buttonlab can improve from 0.42 to 0.80 compared to Deeplab v3+ with traditional augmentation.

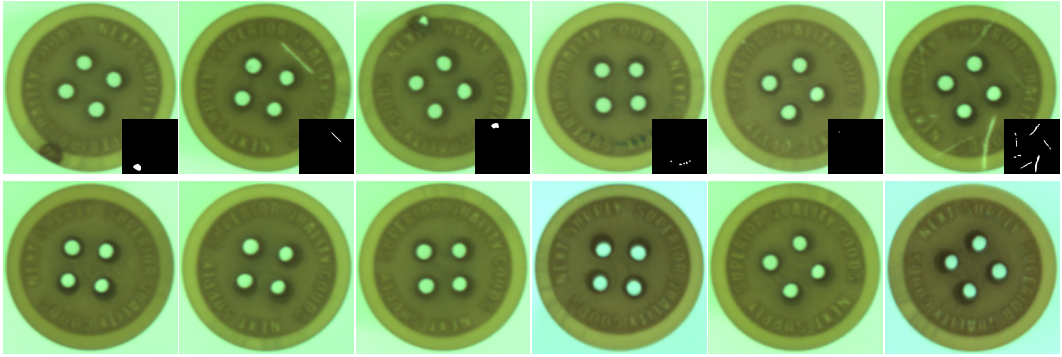

Figure 1. Real samples in the button dataset. The first row is real defect sample with their label, and the second row is some of the defect-free samples.

## 2. RELATED WORK

### 2.1 Neural Style Transfer in the specific area

Gatys et al.[7] proposed Neural Style Transfer in CVPR 2016 to transfer the style from style image $S$ onto a content image $C$, resulting in the fusion of $S$-style and $C$-content. The final output image is obtained by iteratively updating the random-noise image $I$ until the style and content loss are simultaneously minimized.

On this basis, Johnson et al.[8] presented fast forward style transfer and perceptual loss. They used the pre-trained VGG-16 to calculate the loss function of a single style image with a set of content images. The style transfer network can replace the iterative process in Gatys' algorithm[7] to achieve fast style transfer after 2 epochs training. The result is similarly remarkable to Gatys' algorithm[7], but the speed is increased by three orders of magnitude. It takes only 0.015s (on single GTX Titan X GPU) to generate one transferred image (256×256), which can quickly transfer different content images into the same style.

Apart from these global stylization algorithms, Luan et al.[9] designed a local style transfer algorithm in 2018, focusing on the simulation of the local area in the painting. The algorithm consisted of a two-step iteration, which used the patch-match method to find the patch similar to the C image area in the style image for replacement, as well as applied four loss functions to constrain the training. The runtime for a single image (500×500) with 1000 iterations takes 5 mins (on a PC with an Intel Xeon E5-2686 v4 and an NVIDIA Tesla K80 GPU).

The above methods focus on the style transfer of the overall picture or high-quality local picture fusion, which does not fully satisfy the need for rapid generation of local defects. Combining the advantages of the above methods, we design a new generation algorithm, which breaks through the bottleneck of the generation speed as well as ensures the quality of local defect transfer.

### 2.2 Defect Segment Net

Image segmentation is a long-term topic in image processing. Seeking for better segmentation effects, the previous methods often require a small amount of user interaction information as an aid. Grab cut[10] provides two ways of interaction: one uses a bounding box as auxiliary information and the other uses a scribbled line to mark the main object/foreground. After manual labeling, the algorithm can segment the foreground by iterative image division. Even for the image with some complicated background, Grab cut still has a good performance. However, the shortcomings are also very obvious. First, it can deal with semantic segmentation tasks with only one main object. Second, it requires human intervention and cannot be fully automated.

In CVPR 2015, UC Berkeley's Trevor et al.[11] published FCN network, which completed semantic segmentation tasks by training an end-to-end full convolutional neural network. However, the segmentation accuracy is reduced due to pooling and interpolation operations. On this basis, many other segmentation networks have been developed, such as U-Net[12], SegNet[13], PSPNet[14], and Deeplab[15–17], all of which use the structure of Encoder-Decoder. These methods use some approaches to maintain detail information as well as ensure the size of the receptive field.

Chen et al.[18] proposed Encoder-Decoder structure segmentation net Deeplab V3+, which is mainly composed of Deeplab v3[16] as the encoder, while the decoder uses a simple but effective module. The application of improved Xception and atrous Conv can further improve the performance of semantic segmentation tasks. The experimental results show that Deeplab V3+ has achieved new state-of-the-art performance on multiple segmentation task datasets.

Without exception, all segmentation networks require a large number of samples for training. Compared to the semantic segmentation task, application scenarios for defect segmentation task is characterized by simpler foreground information and more monotonous background. Therefore, it is difficult to obtain good results by simply using the semantic segmentation algorithm for defect segmentation. Thus, we add a branch network on the backbone of Deeplab V3+, as well as add local features into the decoder. So that the algorithm can yield better performance when tested on real defect samples, and is more suitable for defect segmentation tasks.

## 3. MATERIAL AND METHODS

The work of this paper is divided into two parts: the generation part and the segmentation part. In the first part, we generate defects on the specified area of the defect-free samples and outputs the corresponding labels of the area. The second part inputs the generated defect samples into the segmentation network for training, and finally obtains the trained segmentation network. The flow of the overall algorithm is shown in **Algorithm 1**.

---

**Algorithm 1:** Automatic Surface Defect Segment $(S, H, M)$

| | |
|---|---|
| **Input** | Defect-free samples as style images $S$ |
| | Several defect samples as hist images $H$, and their masks of the defective area $M$ |
| **Intermediate** | Simulated defect images $\hat{y}$, and their labels of the simulated defective area $l$ |
| **Output** | Segment predicted label $p$ |

**Defect Style Transfer DST:** Trained with a large number of defect-free samples and a single defect sample, the fast feedforward defect simulation network can generate the same type of defect as the reference defect image. Each type needs to train its own DST respectively.

// Use the trained network to quickly generate large quantities of simulated defect images with their labels of the defective area. (§3.1)

$$\hat{y}, l \leftarrow \text{DefectStyleTransfer}(S, H, M)$$

**Defect Segmentation Buttonlab:** The segmentation network is trained with simulation samples of different types of defect. So that we can obtain a well-trained segmentation network capable for detecting all defects. (§3.2)

$$p \leftarrow \text{Buttonlab}(\hat{y}, l)$$

---

## 3.1 Defect Style Transfer

The goal of the simulation algorithm is to generate a large number of defect samples quickly and efficiently. We adopt the idea of two-step method from Luan[9] and replace the first coarse harmonization process with histogram matching, as well as apply the form of Inverse Network from Johnson[8] to accelerate the iterative process in the second step. The defect simulation algorithm flow is as follows.

---

**Algorithm 2**: Defect Style Transfer $(S, L, H, M)$

---

| | |
|---|---|
| **Input** | Defect-free sample $S$ as style image |
| | One single defect sample $H$ as the hist image, and its mask of the defective area $M$ |
| | Area of defect-free sample to generate defect $L$ (it's also the output label $L$) |
| **Output** | Simulated defect image $\hat{y}$, and their label of the simulated defective area $L$ |

**First Step:** Coarse harmonization using hist match in $M$ to maintain color consistency.

// Match the $L$ region of the defect-free image $S$ with the $M$ region of the hist reference sample $H$. The other area remains unchanged.

$$y \leftarrow S[L'] + \text{histmatch}(S[L], H[M])*$$

**Second Step:** Neural style transfer for defect harmonization to maintain texture consistency.

// Transfer Net: Only the specified area of $S$ will be transferred. The Transfer Net's parameters are updated every iteration.

$$\hat{y} \leftarrow S[L'] + \text{TransferNet}(y)[L]$$

// Loss Calculate Net $\phi$: Extracting features with a network pre-trained by ImageNet. The matched image $y$ is taken as the content and the iterative image $C$. The defect-free image $S$ is taken as the style image, and the real defect image $H$ is taken as the hist reference image. Then compare the extracted features to obtain different losses to update the Transfer Net's parameters.

$$\text{Losses} \leftarrow \phi(\hat{y}[L], S[L], H[M])$$

*Annotation: $[\cdot]$ represents the specified area of the image, and $\cdot'$ represents the outside of the area.

---

1   **Hist Match.** The first step of the simulation is to match the specified area with the reference area by histogram matching[19]. The algorithm extracts the histogram information in each channel of defect area of the reference image and then calculates its own histogram equalization map $G_H$. The same operation will be applied to the specified area of the defect-free image to get $G_S$. Finally, the matching image is obtained after $G_H^{-1}(G_S(S[L]))$. Thus, the masked histogram matching module can maintain color consistency of the generated area and the true defect.

2   **Transfer Net.** Transfer Net uses the Unet[12] structure. The encoder part contains three downsampled convolution blocks with four residual blocks behind. The decoder contains two upsampled convolution blocks, and the last jumper connection layer is twice convolved and then passed through the activation layer. Ahead of each convolution, the input feature is expanded with reflection padding to maintain the shape of the output unchanged. To prevent overfitting, we use Instance Normalization[20] after each convolution, followed by a ReLU activation function. Before the network output, we use the Tanh as activation function and use a gaussian mask to fuse the specific defect with the background.

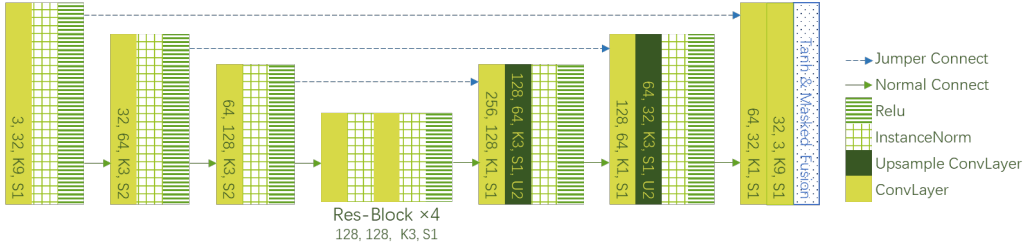

Figure 2. The structure of Transfer Net. The '3, 32, K9, S1' respectively represent the input channel is 3, the output channel is 32, the kernel size is 9 and the stride is 1.

The network input is a masked histogram-matched image, and the output is an image that has been blended into the background after style transfer. The following several losses are applied during training.

**Content Loss and Style Loss.** Johnson et al.[8] adopted two losses in fast style transfer, which were content loss and style loss. Based on this idea, we used pre-train VGG-19 $\phi$ to extract image features. $\phi_j(\cdot)$ represent the $j$-th convolution layer, which is in shape of $C_j \times H_j \times W_j$.

$$\mathcal{L}_{\text{content}}^{\phi,j}(\hat{y},y) = \frac{1}{C_j H_j W_j} \|\phi_j(\hat{y}) - \phi_j(y)\|_2^2 \tag{1}$$

$$\mathcal{L}_{\text{style}}^{\phi,j}(\hat{y},y) = \|G_j^\phi(\hat{y}) - G_j^\phi(y)\|_F^2 \tag{2}$$

Where the gram matrix is $G_j^\phi(x)_{c,c'} = \frac{1}{C_j H_j W_j} \sum_{h=1}^{H_j} \sum_{w=1}^{W_j} \phi_j(x)_{h,w,c} \phi_j(x)_{h,w,c'} = \psi\psi^T / C_j H_j W_j$, $\psi$ is the $\phi_j(\cdot)$ reshaped features of $\phi_j(\cdot)$ in shape of $C_j \times H_j W_j$.

Thus, we can get weighted Gatys losses as below. Where $w_s$ and $w_c$ is the weight of two losses.

$$\mathcal{L}_{\text{Gatys}} = w_c \mathcal{L}_c + w_s \mathcal{L}_s \tag{3}$$

By using these two losses, the generated image can be blended into the texture features of the style image while being perceptually similar to the content. Different style transfer results are controlled by applying features $\phi_j(\cdot)$ from different layers.

**Hist Loss.** Simply calculating the above two kinds of losses will result in the lack of texture consistency and blur in generated images. So we employ Risser's[19] histogram loss.

$$\mathcal{L}_{\text{hist}} = \frac{1}{C_j H_j W_j} \|\phi_j(\hat{y}) - R_j(\hat{y})\|_F^2 \tag{4}$$

Where $R_j(\hat{y}) = \text{histmatch}(\phi_j(\hat{y}), \phi_j(y_{\text{hist}}))$, represents the result of histogram matched $\phi_j(\hat{y})$ with the reference image in each channel.

Luan[9] use style images as a reference for hist loss. Considering that only a few real samples in the defect simulation can be utilized, for each type of defect, we select the real sample as the hist reference image $H$.

**Total Variation Loss.** Using tv loss can increase the smoothness of the transitions between specified transferred region and background[8,21], making the composite region less abrupt.

$$\mathcal{L}_{\text{tv}}(\hat{y}) = \sum_{x,y} (\hat{y}_{x,y} - \hat{y}_{x,y-1})^2 + (\hat{y}_{x,y} - \hat{y}_{x-1,y})^2 \tag{5}$$

In summary, the weighted loss function is as below. Where $w_h$ and $w_{tv}$ are the weight of the hist loss and the tv loss.

$$\mathcal{L}_{whole}(\hat{y}) = \mathcal{L}_{Gatys}(\hat{y}[L], C[L], S[L]) + w_h \mathcal{L}_{hist}(\hat{y}[L], H[M]) + w_{tv} \mathcal{L}_{tv}(\hat{y}[L]) \quad (6)$$

After a forward propagation through the trained network, we can blend the rough matching image into the background to obtain the defect simulation samples. Part of the simulation samples and their intermediate hist match outputs are shown in Figure 3.

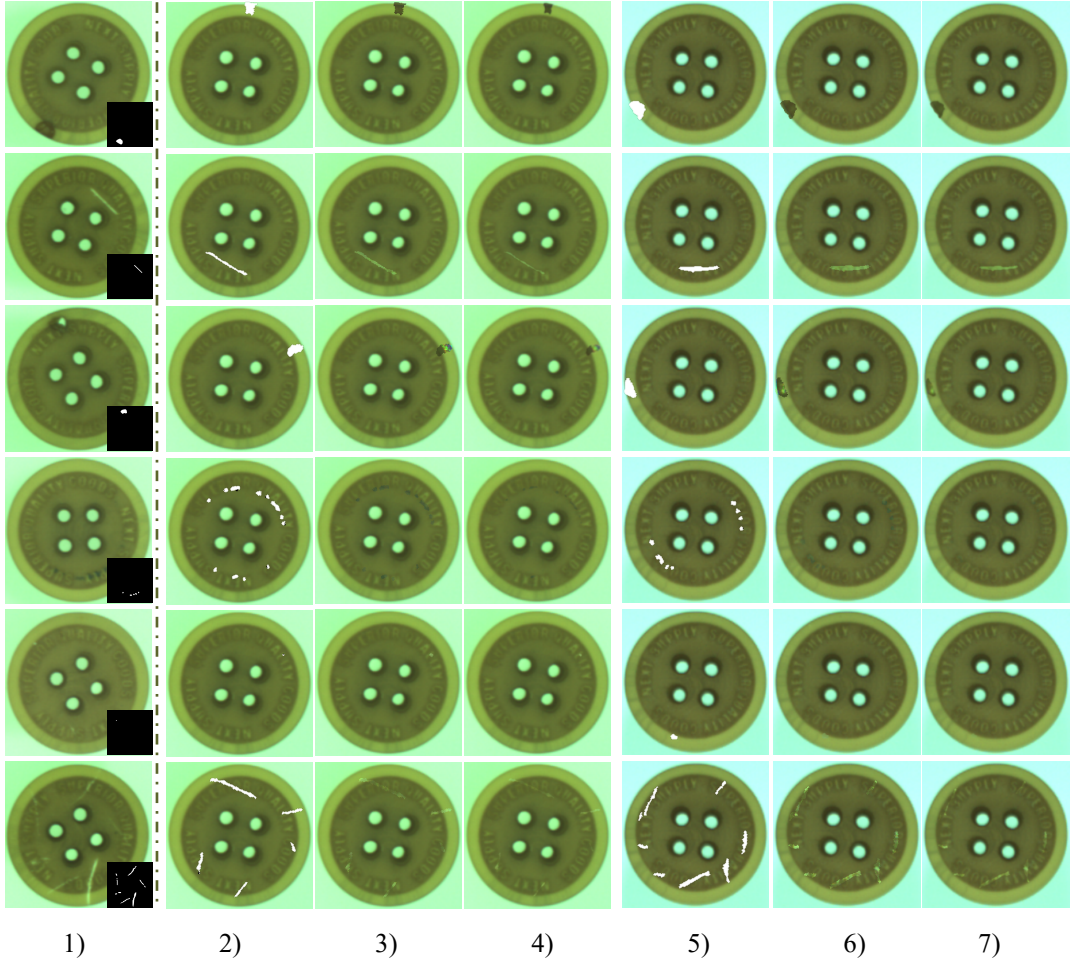

1)    2)    3)    4)    5)    6)    7)

Figure 3. The results of generating different defects in the same sample. The 1st column is the real defect reference samples. The 2nd and 5th columns are defect-free samples with different masks. The 3rd and 6th columns are the output of Hist Match. The 4th and 7th columns are the output of the whole proposed DST. Each row represents one type of defect.

## 3.2 Segment Net Buttonlab

In the surface defect detection task, the segmentation of small-scale and weak defects requires more use of low-level local features. In general, the lower the level of feature, the more complete the local information is retained. Therefore, we add a branched net based on Deeplab v3+ to extract fine local features. The branch net consists of a 7×7 convolution and the first residual block structure in ResNet-50[22]. After the branch net, the image resolution is reduced to 1/2, and the local feature is well preserved. In order to fuse the features extracted by the branch network, we modify the decoder network to merge the global low-level and high-level features of the backbone net and the local features of the branch network. Then we restore the resolution gradually by three times of sampling. Considering that the semantic information in the defect segmentation task is simpler and the size of the training sample is smaller, we select the lightweight ResNet-50 as the backbone network. In addition, because the scale of the same kind of defects is relatively fixed, we remove the ASPP pooling layer which is meant for the multi-scale feature. The structure of Buttonlab is shown in Figure 4.

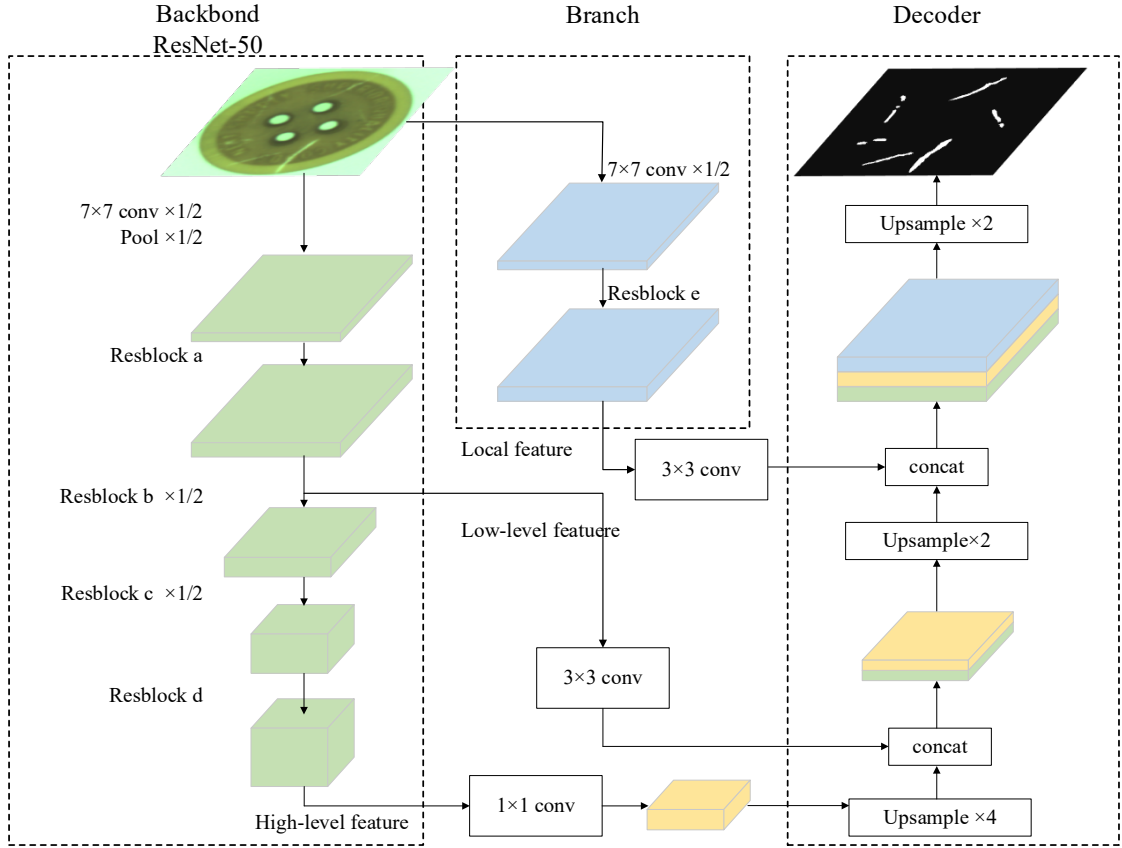

Figure 4. The structure of Buttonlab. We add branch net to extract local feature. Together with global low-level and high-level features, the whole segment net is more suitable for defect segmentation tasks than Deeplab v3+.

In order to increase the network randomness during the training process and make the branch network better extract the local features, we propose a partial image input strategy, as shown in Figure 5. During training, the input of the backbone network is the whole image, while the input of the branch network is an image block randomly cropped from the whole image. At this point, the output of the network will no longer be a full-size segmented image, but a predicted block of the same size as the input block. During the backward, we only calculate the loss inside the cropped image block.

This strategy has the following advantages of three perspectives. First, the branch net is forced to focus on local features, since the input is a partial image, from which global information is unable to be obtained. However, according to the global information provided by the backbone network, the decoder can still make sense to obey the global semantics. Second, the generalization ability of the network can be improved by increasing the randomness of the input of branch net. In the absence of the training samples, the random cropping strategy is equivalent to implicitly augmenting the dataset, thereby reducing the model overfitting. Third, the computational cost of the branch net during training is reduced, improving the training efficiency.

We use masked cross-entropy as the loss function, and the optimization goal of Buttonlab is shown in (7). Where $p_i[\text{crop}]$ and $l_i[\text{crop}]$ represent the prediction and label of the $i$-th image block. The algorithm flow of the training process is shown in Algorithm 3.

$$\mathcal{L}_{seg} = \sum_{i=1}^{N} \mathcal{L}_{ce}\left(p_i[\text{crop}], l_i[\text{crop}]\right) \tag{7}$$

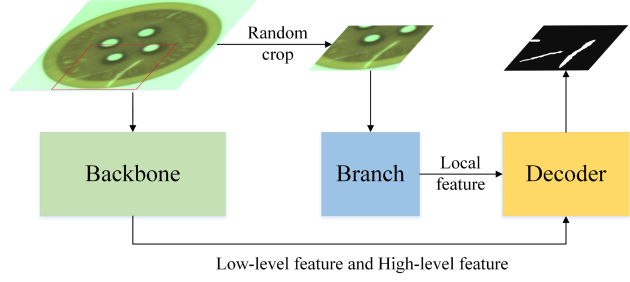

Figure 5. Schematic diagram of random cropping strategy during training, making the branch net focus on local features. The strategy will not be used during validation or test.

---

**Algorithm 3:** Segment Net Buttonlab ( $y$ , $l$ )

---

**Input**       Simulated and real defect images $y$, and their labels $l$

**Output**      Segment predicted label $p$

**Train:** Random crop the input image and label to obtain an image block $y[\text{crop}]$ and its label block $l[\text{crop}]$.

$$(y[\text{crop}], l[\text{crop}]) \leftarrow \text{RandomCrop}(y, l)$$

// Forward: Enter the full image $y$ and the cropped image block $y[\text{crop}]$ respectively into Backbone and Branch to get the predicted block label $p[\text{crop}]$.

$$p[\text{crop}] \leftarrow \text{Decoder}\big(\text{Backbone}(y), \text{Branch}(y[\text{crop}])\big)$$

// Backward: Calculate the loss $\mathcal{L}_{seg}$ between $p[\text{crop}]$ and $l[\text{crop}]$, and then use the Adam optimizer to update network parameters.

$$\text{Loss} \leftarrow \mathcal{L}_{seg}\big(p[\text{crop}], l[\text{crop}]\big)$$

**Test:** Enter the whole images simultaneously into Backbone and Branch to get the predicted label $p$.

$$p \leftarrow \text{Decoder}\big(\text{Backbone}(y), \text{Branch}(y)\big)$$

---

## 4. EXPERIMENT

In this chapter, we set up two sets of experiments. The first set is a comparison experiment to verify two aspects: validity of the generated algorithm DST, as well as the superiority of the Buttonlab in defect segmentation task. We compare the performance of different combinations of simulation and segmentation algorithms. The second set is to evaluate the effect of the simulated defect sample on the Buttonlab. We compare the segmentation accuracy of the Buttonlab with and without the addition of simulation samples when using different numbers of real defect samples.

The experiments are performed on the button dataset. We select 9 real defect samples with different defects as reference samples for simulation algorithms and 238 real defect-free samples as background. Each simulation algorithm generates 2167 simulation samples for training segment algorithms. Another 78 real defect samples are selected as test sets. Besides, there are 150 samples with real defects prepared for the second experiment.

The algorithm is implemented on the PyTorch[23]. We use Adam[24] optimizer for DST and set the learning rate 1e-3. We train 10 epochs, and the weights in (6) are $w_c = 1e5$, $w_s = 2.5e-4$, $0.1w_h = w_{tv} = 1$. We use Adam optimizer with learning rate 2e-3 for Buttonlab. It takes 5 mins to train a single defect DST on intel i5-8500 + Nvidia 1070Ti 8G. Trained DST can generate up to 20 images per second.

The experiment uses the F1 score as an evaluation index, and its definition is as shown in (8).

$$F1 = \frac{2 \cdot precision \cdot recall}{precision + recall} \tag{8}$$

Where $precision = \frac{TP}{TP+FP}$ and $recall = \frac{TP}{TP+FN}$.

### 4.1 Comparison experiment

The comparison experiment mainly evaluates the F1 scores of different combinations of generation and segmentation networks on the same test set. We use the proposed DST, hist match and cycleGAN[25] to generated defect sample. In all cases, the reference defect samples are exactly the same 9 real images and the simulated defects locate in the same area of the same 238 defect-free samples; each generation algorithm produces 2167 images. Besides, we also set a control group of using only 9 real samples for training the segment net. We use Proposed Buttonlab, Deeplab v3+[18], and U-Net[12] as defect segment net. Each algorithm trains 120 epochs with batch size 8 over the same train dataset, all the images randomly sampled without replacement in each epoch.

Table 1. Results of Comparison Experiment. Each grid represents a combination of different simulation algorithms with different segmentation algorithms.

| Segmentation | Simulation | | | Only real |
| --- | --- | --- | --- | --- |
| | Proposed DST | Hist match | cycleGAN | |
| Proposed Buttonlab | **0.8000** | **0.6599** | **0.6841** | 0.4692 |
| Deeplab v3+ | 0.6390 | 0.4752 | 0.5693 | 0.4249 |
| U-Net | 0.6578 | 0.1930 | 0.6377 | **0.5903** |

Table 1 shows the result of the comparison experiment. By comparison, among the three simulation algorithms, DST produces the highest quality simulated samples for training segment algorithms. The proposed Buttonlab segmentation algorithm achieves the optimal segmentation accuracy with each type of simulation samples. Compared to using only a small number of real defect samples for training, the additional generated simulation samples can significantly improve the segmentation accuracy of the segmentation models.

These results suggest that DST generated defect samples can significantly improve the performance of the segmentation algorithm with scarce samples, and the proposed Buttonlab is more suitable for defect segmentation task.

### 4.2 Influence of simulated defect samples on Buttonlab

The second experiment mainly verifies the effect of the addition of simulated defect samples on the training of Buttonlab. We compare the F1 score of Buttonlab on the same test set with and without the addition of simulation samples. 9 real defect samples are used as reference samples $H$ for the training simulation network, in addition to adding 50, 100, 150 real defect samples for training Buttonlab. Strategy A is trained with real defect samples and adopts random sampling without replacement. Strategy B is only trained with real defect samples. Strategy C is trained with both simulation samples and real defect samples. Both B and C are trained 100 batches in each epoch by random sampling with replacement, with a batch size of 8. Each strategy trains 120 epochs.

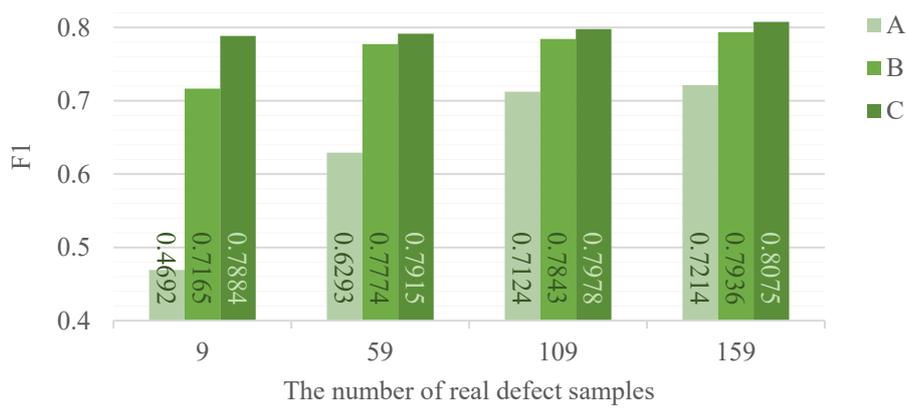

Figure 6. Analysis of the number of real defect samples with and without the simulation samples. Trained samples of Strategy A are sampled without replacement, while B's and C's are sampled with replacement and trained 100 batches every epoch. Strategy A and B are only trained with the real images, and C is trained with the real and the generated simulation samples.

Figure 6 shows the results of the experiment. Figure 7 shows the prediction of different models trained with different samples. It's obvious that the use of simulation samples can significantly improve the performance of the defect segmentation net, and is particularly effective in the case of insufficient real defect samples. In view of the scarcity of the actual production defect sample and the high labeling cost, using the proposed DST with Buttonlab can significantly reduce the demand of the defect samples for training a neural net, which only uses one real defect sample for each type of defect. Besides, when the samples are sufficient, the simulation samples can further improve the performance of the segmentation network.

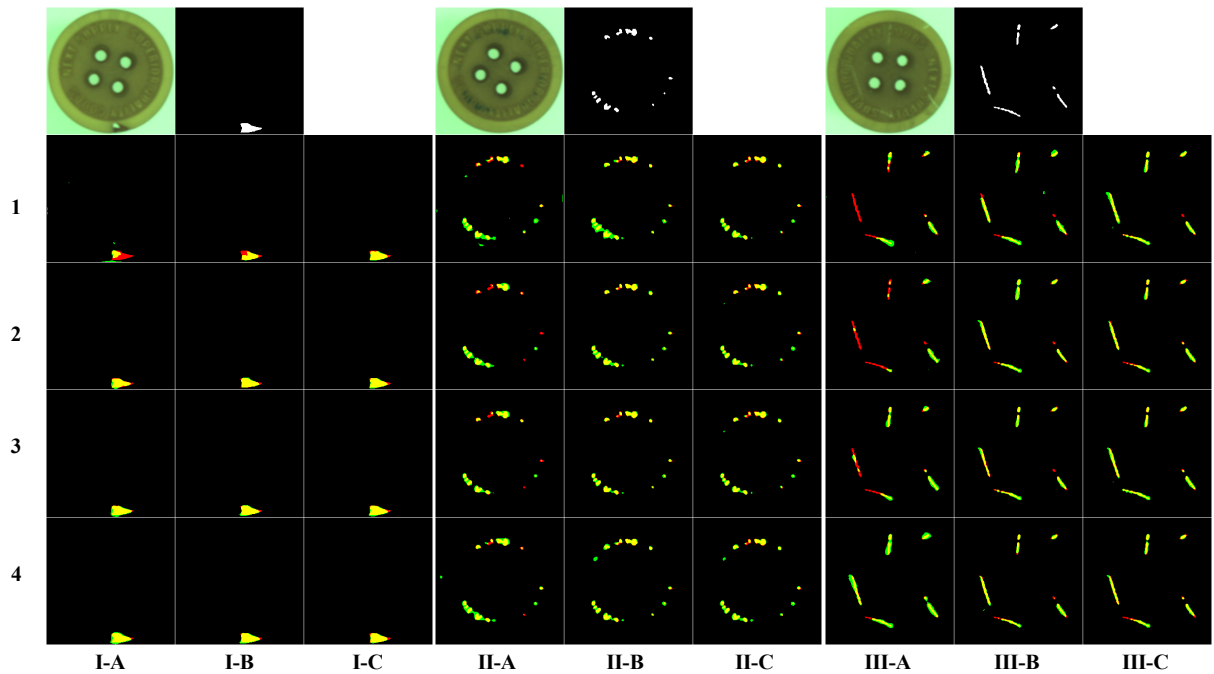

Figure 7. Eval of models trained with different samples. Ⅰ-Ⅲ represent different samples. 1-4 represent the number of trained samples are 9, 59, 109, 159. A-C indicates different strategies. Red represents the manual label, while green represents the prediction of different models. Yellow represents the correct segment area.

## 5. CONCLUSION

In order to solve the contradiction between the difficulty in obtaining the defect samples in actual production and the requirement of a large number of labeled samples for training a neural network, we propose a method to generate simulated defect samples by using style transfer algorithm. The proposed DST only uses a single labeled real defect sample as the reference sample together with some real defect-free samples for training. A trained DST can quickly generate a large number of defect samples with the same type of defect as the reference within a specified area of the defect-free images. According to the characteristics of defect segmentation task, we also design Buttonlab for defect segmentation task on the basis of Deeplab v3+. On the button dataset, combined with the two proposed algorithms, our approach can achieve the F1 score of 0.8000 using only 9 real defect samples. Compared with other generation algorithms and segmentation algorithms, the proposed algorithms can get better results in the case of insufficient samples. And when the sample is sufficient, the generated defect samples can further improve the segmentation accuracy.

If the difference between the defect area and the background is small, it is difficult for the segmentation network to recognize it. How to make good use of all the information generated by the sample, as well as improve the segmentation accuracy is the research goal of our next work.